\documentclass[runningheads]{llncs}
\usepackage[T1]{fontenc}
\usepackage{graphicx}
\usepackage{booktabs}
\usepackage[misc]{ifsym}
\usepackage{amsmath} 
\usepackage{amssymb}
\usepackage{multirow}
\usepackage{subfigure}
\usepackage{makecell}

\usepackage{mwe}
\usepackage{ulem}

\begin{document}

\title{I-GLIDE: Input Groups for Latent Health Indicators in Degradation Estimation}

\titlerunning{Input Groups for Latent Health Indicator in Degradation Estimation} 


\author{Lucas Thil\inst{1, 2} \Letter \and Jesse Read.\inst{1} \and Rim Kaddah\inst{2} \and Guillaume Doquet\inst{3}}

\authorrunning{Thil et al.}
\toctitle{I-GLIDE: Input Groups for Latent Health Indicators in Degradation Estimation}
\tocauthor{Lucas Thil, Jesse Read, Rim Kaddah, Guillaume Doquet}

\institute{LIX Ecole Polytechnique, France \email{thil@lix.polytechnique.fr} \email{jesse.read@polytechnique.edu} \and IRT SystemX, France \email{lucas.thil@irt-system.fr} \email{rim.kaddah@irt-systemx.fr} \and Safran Tech \email{guillaume.doquet@safrangroup.com}}

\maketitle              

\begin{abstract}
Accurate remaining useful life (RUL) prediction hinges on the quality of health indicators (HIs), yet existing methods often fail to disentangle complex degradation mechanisms in multi-sensor systems or quantify uncertainty in HI reliability. This paper introduces a novel framework for HI construction, advancing three key contributions. First, we adapt Reconstruction along Projected Pathways (RaPP) as a health indicator (HI) for RUL prediction for the first time, showing that it outperforms traditional reconstruction error metrics. Second, we show that augmenting RaPP-derived HIs with aleatoric and epistemic uncertainty quantification (UQ)—via Monte Carlo dropout and probabilistic latent spaces— significantly improves RUL-prediction robustness. Third, and most critically, we propose indicator groups, a paradigm that isolates sensor subsets to model system-specific degradations, giving rise to our novel method, I-GLIDE which enables interpretable, mechanism-specific diagnostics. Evaluated on data sourced from aerospace and manufacturing systems, our approach achieves marked improvements in accuracy and generalizability compared to state-of-the-art HI methods while providing actionable insights into system failure pathways. This work bridges the gap between anomaly detection and prognostics, offering a principled framework for uncertainty-aware degradation modeling in complex systems.

\end{abstract}

\section{Introduction}


Accurate RUL prediction is critical for enabling condition-based maintenance in complex engineering systems. A cornerstone of this task lies in deriving interpretable Health Indicators (HIs) that reliably capture subsystem degradation patterns. While autoencoder (AE)-based reconstruction errors have emerged as a popular HI-construction method, existing approaches suffer from two key limitations: (1) sensitivity to noise and epistemic uncertainty, which obscures degradation signals, and (2) a lack of granularity in disentangling subsystem-specific degradation behaviors. This work addresses these gaps by introducing a novel Ensemble of Indicators framework, which advances traditional AE architectures through multi-head encoders and decoders designed to isolate degradation patterns across subsystems (e.g., fan, high-pressure compressor). We call this method I-GLIDE: Input Groups for Latent Health Indicators in Degradation Estimation.

Contrary to prior studies that treat UQ as an auxiliary feature, our proposed method I-GLIDE leverages this uncertainty to enhance HI robustness while maintaining explainability. 
We rigorously benchmark our approach against established latent-space HI methods—including Reconstruction along Projected Pathways (RaPP) \cite{Ki_Hyun_Sangwoo_Yongsub,Thil} and Monte Carlo (MC) dropout-based uncertainty estimation \cite{gal2016dropoutbayesianapproximationrepresenting} demonstrating superior RUL prediction accuracy on the NASA C-MAPSS turbofan dataset \cite{Saxena_Goebel_Simon_Eklund_2008} and the MILL NASA degradation dataset \cite{mill_nasa}. Our contributions are threefold:


\begin{enumerate}
    \item \textbf{Systematic Analysis}: We identify and characterize critical limitations of existing AE derived HIs, notably their vulnerability to noise and inability to isolate subsystem-level degradation.
    \item \textbf{Uncertainty-Aware Benchmarking}: By integrating aleatoric and epistemic UQ into latent-HI construction, we improve RUL estimation.
    \item \textbf{I-GLIDE Framework}: We propose a multi-head AE architecture where each encoder-decoder pair targets distinct subsystems, enabling granular, explainable HI extraction, achieving state-of-the-art RUL prediction while providing insights into degradation mechanisms. 
\end{enumerate}

\begin{table}[h!]
\centering
\caption{Notation used in the paper.}
\begin{tabular}{ll}
\hline
\textbf{Symbol} & \textbf{Description} \\ \hline
$\sigma_a, \sigma_e$ & Aleatoric, Epistemic uncertainty \\ 
$\mathcal{F}$ & Function mapping HIs to a RUL \\ 
$\mathcal{X}$ & Input data set with entries $x$ \\ 
$\hat{x}$ & reconstruction of input $x$, also noted as the target variable $y$ \\ 
$W_{D}$ & Decoder weight parameters \\ 
$g \in G$ & Set of sub-complex systems indices $g$ \\ 
$z$ & Latent space of the AE \\ 
$y$ & Target variable \\ 
$h_{g,l}$ & hidden layer of group $g$ $h_g$, at position $l$ \\ 
$d_g(x)$ & specific distance vector of groups $h_g(x)-h_g(\hat{x})$ \\ \hline 
\end{tabular}
\label{tab:notation}
\end{table}

\section{Background and Related Works}  
\subsection{RUL Prognostics and Health Indicators}

Most industrial complex systems are built by the interdependencies of sub-complex systems; degradation in one component can propagate cascading effects, triggering operational disruptions, escalating costs, safety risks, and—in extreme cases—catastrophic system-wide failures. As a result, the accurate RUL estimation of a complex system is heavily studied in engineering, particularly where costs and safety are associated. Early methodologies relied on stochastic approaches, such as threshold-based degradation signatures or empirical lifetime metrics (e.g., flight cycles or mileage) \cite{Fink_Wang_Svensén_Dersin_Lee_Ducoffe_2020}. While methods prioritized identifying failure precursors or tracking cumulative usage they lacked adaptability to complex, non-linear degradation patterns. Timely and precise RUL prognostics not only curtails downtime and waste but also enables proactive maintenance, aligning operational decisions with evolving system health.

\subsection{Evolution of HI Extraction}  
Early HI derivation prioritized interpretability through handcrafted statistical features (e.g., signal variance) or physics-based models. 
Using a a Bayesian framework enriched by expert knowledge to estimate failure probabilities, Lacaille \cite{Lacaille_2009} proposed a normalization pretreatment to derive standardized signatures interpretable as HIs by domain experts. However, such approaches depended heavily on predefined failure patterns and manual refinement, limiting their adaptability to heterogeneous operational conditions in non-stationary environments \cite{come_etienne}. Hybrid approaches combined Kalman filters with NNs to model state-of-charge degradation \cite{He_Williard_Chen_Pecht_2014}, while others used neural networks (NNs) to learn a RUL representation to derive syncretic HIs \cite{Wei_Ye_Wang_Xinxin-Xu_Twajamahoro_2022}. Zhao et al. \cite{Zhao_Bin_Liang_Wang_Lu_2017} used degradation pattern learning in the case of turbofan engines to predict the RUL. They extracted degradation patterns that helped characterize the nature of the degradation, which can itself be seen as a HI. Furthermore, their method was shown to improve the predictive capability of a NN towards RUL estimation. 

AEs later emerged as a cornerstone method, using reconstruction errors from healthy-state training as implicit HIs \cite{Huang_Pan_Liu_Gong_2023,Martinelli_Tronci_Dipoppa_Balducelli_2004,pmlr-v101-gherbi19a}. Despite progress, these methods often assumed linear degradation trends or predefined failure modes, limiting adaptability to non-stationary systems. Other NN approaches later enabled data-driven prognostics, with Long Short-Term Memory (LSTM) architectures capturing temporal degradation in batteries \cite{Wang_Fan_Jin_Takyi-Aninakwa_Fernandez_2023} and turbofans \cite{Li_Ding_Sun_2018}. However, early frameworks often bundled HI estimation with RUL prediction, risking conflated objectives where HIs were implicitly tuned to downstream tasks rather than intrinsic degradation patterns. This coupling became particularly evident in methods that embedded domain assumptions directly into HI design. For example, Jing et al. \cite{Jing_Zheng_Xia_Liu_2022} incorporated exponential normalization of sensor data as an inductive bias in a Variational Autoencoder (VAE), aligning the HI with the CMAPSS dataset’s predefined degradation trends. While this yielded robust RUL predictions, it effectively tied the HI to the target degradation profile, limiting adaptability to systems with non-exponential behaviors. Pillai and Vadakkepat \cite{Pillai_Vadakkepat_2021} addressed this issue more directly, proposing a two-stage architecture that decoupled HI feature discovery from RUL regression.
Their approach improved generalizability by isolating degradation modeling from task-specific optimization, though challenges persisted in interpretability and subsystem-specific analysis.

\subsubsection{Latent-Space HI Refinement}


More recent advances focused on refining HI quality through latent-space analysis. Kim et al. \cite{Ki_Hyun_Sangwoo_Yongsub} introduced the RaPP method, projecting latent representations from an AE's encoder to compute distance metrics that outperformed classical reconstruction errors. González-Muñiz et al. \cite{González-Muñiz_Díaz_Cuadrado_García-Pérez_2022} validated this paradigm shift, demonstrating that latent-space metrics from RaPP consistently surpass input-space approaches in HI quality. Their work highlighted the latent space between encoder and decoder as a rich source of degradation signals, though subsystem-specific trends remained obscured by holistic aggregation. Despite these innovations, mapping HIs to RUL remains fraught with challenges. Many approaches employ black-box models or simplistic linear mappings \cite{Datong_Liu_Jianbao_Zhou_Haitao_Liao_Yu_Peng_Xiyuan_Peng_2015}, neglecting context-dependent HI interpretations under varying operational conditions. For instance, a high reconstruction error might indicate severe degradation in one context but sensor noise in another—a nuance often lost in end-to-end frameworks. Recent benchmarking by Rombach et al. \cite{Rombach_Michau_Bürzle_Koller_Fink_2024} underscores this gap, advocating for feature engineering to improve HI interpretability while maintaining correlation with ground truth degradation.

\subsection{Uncertainty-Aware Subsystem Modeling}  
Uncertainty quantification in NNs can be achieved through MC dropout \cite{Abdar_Pourpanah_Hussain_Rezazadegan_Liu_Ghavamzadeh_Fieguth_Cao_Khosravi_Acharya_et_al_2021}. Variational AEs (VAEs) \cite{Akbari_Jafari_2020}, can disentangle aleatoric uncertainty (inherent data noise) and epistemic uncertainty (model ambiguity) to isolate distinct sources of unpredictability.
While probabilistic frameworks optimize maintenance via confidence intervals \cite{Mitici_De_Pater_Barros_Zeng_2023}, UQ is often treated as a post-hoc refinement rather than a core HI component. Deterministic AEs, for instance, cannot isolate aleatoric uncertainty due to fixed latent spaces—a limitation addressed by variational architectures \cite{Wang_Pang_Chen_Iyer_Dutta_Menon_Liu_2021}. Ensemble methods further reduce uncertainty \cite{Zhang_Lim_Qin_Tan_2017}, yet their application to subsystem-aware HIs remains underexplored.  

\subsection{Monotonicity and Degradation Dynamics}  
RUL is typically modeled as a monotonic function of the State of Health (SOH), declining from 100\% (pristine) to 0\% (failure). While mechanical wear rarely reverses, subsystem interactions (e.g., turbine degradation accelerating fan wear) introduce non-stationary dynamics \cite{Amini_Soleimany_Karaman_Rus_2018}. This necessitates HIs that isolate localized degradation while preserving system-wide coherence—a gap addressed by our subsystem-aware architecture. 
Similarly, we model RUL estimation as a function $\mathcal{F}$ of the HIs, mapping their values to the corresponding RUL.

\section{Proposed Approach: I-GLIDE}


In order to produce better HIs, we build on foundational assumptions about degradation dynamics and their statistical relationships to RUL established earlier, and we begin by formalizing our UQ for prognostic tasks. We then introduce a novel architecture that disentangles subsystem-specific degradation signatures by managing sensor groups and operational variabilities at the component level, while adapting the RaPP method to mitigate cross-component interference. 
Finally, we propose a data-driven strategy to validate constructed HIs through direct RUL estimation, demonstrating their prognostic utility. Each phase is rigorously evaluated via empirical case studies (Section 4), ensuring robustness across diverse degradation scenarios. Our notation is summarized in Table \ref{tab:notation}.

\subsection{Uncertainty Quantification}  
\label{subsec:uncertainty}  

Uncertainty in prognostics arises from two primary sources: aleatoric ($\sigma_a$), inherent to data noise and irreducible even with additional observations, and epistemic ($\sigma_e$), stemming from model limitations and reducible through improved architectures or training \cite{Hüllermeier_Waegeman_2019,Kendall_Gal_2017}.

In order to produce high-quality HIs we make use of the UQ capabilities of AE architectures with an underlying change. Our epistemic UQ focuses on the scalar reconstruction error $\epsilon = \|x - \hat{x}\|_2$ instead of the full-dimensional decoder output $\hat{x}=y$. This aligns with prognostics frameworks where $\epsilon$ serves as a health indicator (HI), reducing dimensionality for easier integration with downstream RUL prediction models (e.g., $\mathcal{F}$). We avoided using raw $y$ (the reconstruction) as a standalone HI directly because it is 1) outperformed by RaPP methods \cite{Ki_Hyun_Sangwoo_Yongsub,González-Muñiz_Díaz_Cuadrado_García-Pérez_2022} and 2) our RUL predictor model $\mathcal{F}$ showed high variance in selecting the best variables when both RaPP and $y$ were fed as inputs. Thus we dropped $y$ as a HI and instead focused in introducing MC dropout to quantify $\epsilon$ uncertainties as a HI which showed to be a better complement.
Therefore, the disentangled UQ is performed through the aggregation of our $\epsilon_1..\epsilon_n$ over $n$ MC samples in our VAE architecture:

\[
\sigma_a = \text{Var}\left(\epsilon_1, \dots, \epsilon_n\right|\text{fixed } \text{W}_\text{D}), \quad 
\sigma_e = \text{Var}\left(\epsilon_1, \dots, \epsilon_n\right|\text{fixed } z).
\]  

As \( z \) is deterministic, aleatoric uncertainty \( \sigma_a \) cannot be isolated in AEs, rendering it undefined. Thus, we can only compute $\sigma_e$ in the case of a vanilla AE. This highlights the advantage of VAEs for joint uncertainty estimation.

\subsection{Architecture}

\begin{figure}[t]
\label{fig:ensemble_architecture1}
\includegraphics[width=\textwidth]{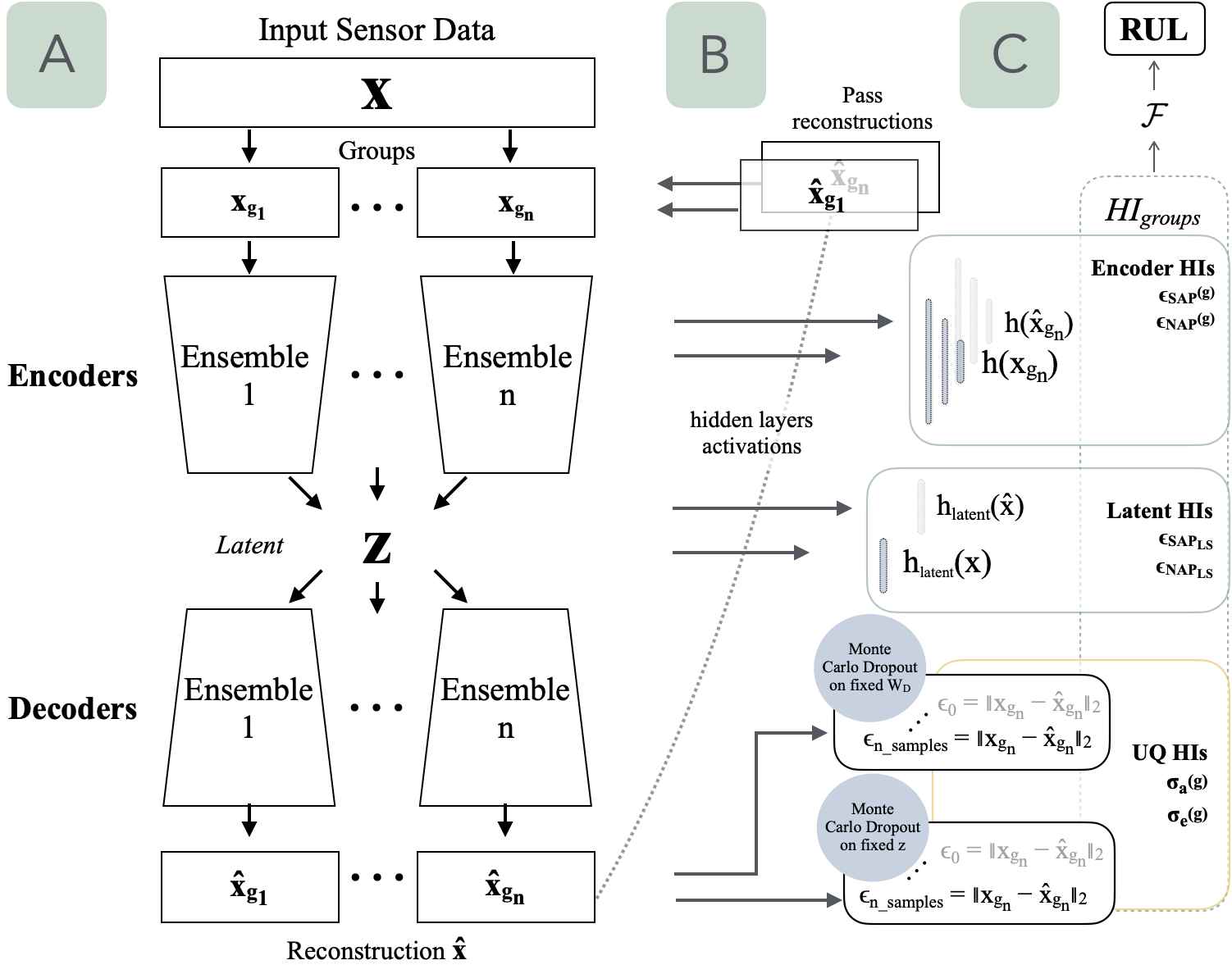}
\caption{\textbf{I-GLIDE Architecture Framework} --  
\textbf{A:} Subsystem-specific encoder-decoder heads learn distinct latent representations, fused into a shared latent space \( z \) via reconstruction loss (trained on healthy data).  
\textbf{B:} HIs are extracted using RaPP metrics \cite{González-Muñiz_Díaz_Cuadrado_García-Pérez_2022} and UQ \cite{Kingma_Welling_2013} over full trajectories.  
\textbf{C:} Aggregated HIs are used to predict RUL, trained via a Random Forest (RF) regressor \( \mathcal{F} \).}  

\end{figure}

Building on the above foundations, our proposed architecture extends the traditional AE and VAE frameworks by introducing multiple encoder-decoder pairs for each sensor group, which are then integrated through a shared latent space, as illustrated in Fig. 1 (\ref{fig:ensemble_architecture1}). This design addresses the non-stationarity of sensor signals by disentangling subsystem-specific degradation dynamics in the latent space. This separation allows us to apply the RaPP \cite{Ki_Hyun_Sangwoo_Yongsub} method individually to each encoder. By projecting the activations of the hidden spaces $h_g(x)$ corresponding to isolated sensor groups $g \in G$, we aim to achieve more comprehensive feature extraction, enabling the construction of  specific health indicators (HIs).

We derive this architecture with two different latent spaces: the first one being in the way of traditional AE named $\text{I-GLIDE}_\text{AE}$, and in the second version where the latent is a Gaussian type distribution in the manner of VAEs named $\text{I-GLIDE}_\text{VAE}$. In the latter, we can leverage the variational inference aspect of the architecture.

\subsection{Adapting Domain-Specific Latent Space Health Indicators}  


A main novelty of I-GLIDE is to adapt RaPP \cite{Ki_Hyun_Sangwoo_Yongsub}, traditionally used in monolithic architectures, for each subsystem in our multi-autoencoder framework. By moving away from the monolithic approach, I-GLIDE computes group-specific health indicators (HIs) for each sensor group \( g \). Unlike the original RaPP framework, which operates on a single encoder-decoder pair, our architecture independently calculates \( \epsilon_{\text{SAP}^{(g)}} \) and \( \epsilon_{\text{NAP}^{(g)}} \) for each group \( g \), leveraging dedicated encoders and decoders per subsystem. These components share a cohesive latent space \( z \), preserving global system coherence while isolating localized anomalies. For a sensor group \( g \), let \( h_{g,l}(x) \in \mathbb{R}^{n_l} \) denote the activations of the \( l \)-th encoder layer (with \( n_l \)-dimensional output) for input \( x \), and \( h_{g,l}(\hat{x}_g) \) represent the reconstructed activations across \( L \) layers (\( l=1,\ldots,L \)).  



We redefine two HIs per sensor group:  
\begin{equation}
    \epsilon_{\text{SAP}^{(g)}}(x) = \left\| h_{g}(x) - h_{g}(\hat{x}) \right\|_2 
\end{equation} 
which computes the first RaPP metric: the Simple Aggregation along Pathway (SAP) \cite{Ki_Hyun_Sangwoo_Yongsub} as the Euclidean distance between original and reconstructed activations across all layers \( l \). For the second RaPP metric, Normalized Aggregation along Pathway (NAP), we first derive the group-specific distance vector \( d_g(x) = h_{g}(x) - h_{g}(\hat{x}) \), where \( h_{g}(x) = [h_{g,1}(x), \dots, h_{g,L}(x)] \) concatenates activations across all layers $l$. Given a training set \( \mathcal{X} \), let \( D_g \) be a matrix whose rows correspond to \( d_g(x_i) \) for \( x_i \in \mathcal{X} \), and let \( \bar{D}_g \) denote the column-wise centered version of \( D_g \). The NAP metric for group \( g \) is then:  
\begin{equation}
    \epsilon_{\text{NAP}^{(g)}}(x) = \left\| \left(d_g(x) - \mu_{\mathcal{X}}\right)^\top V_g \Sigma_g^{-1} \right\|_2.
\end{equation}
Here, \( \mu_{\mathcal{X}_g} \in \mathbb{R}^{n_l \cdot L} \) is the column-wise mean of \( D_g \), \( \Sigma_g \in \mathbb{R}^{k \times k} \) is a diagonal matrix containing the singular values of \( \bar{D}_g \), and \( V_g \in \mathbb{R}^{(n_l \cdot L) \times k} \) contains the right singular vectors from the singular value decomposition (SVD) of \( \bar{D}_g \), with \( k \) denoting the rank of \( \bar{D}_g \).  

This design allows the model to isolate sensor group contributions, where anomalies in specific sensor groups are preserved without being diluted by nominal signals from other groups and this results in enhanced interpretability as HIs directly map to physical sensor groups, aiding root-cause analysis.

Building on González et al. \cite{González-Muñiz_Díaz_Cuadrado_García-Pérez_2022}, where latent-space RaPP metrics outperform encoder-derived counterparts for HI construction, our method integrates both approaches. We compute latent-space $z$ metrics ($\epsilon_{\text{NAP}_{\text{LS}}}$, $\epsilon_{\text{SAP}_{\text{LS}}}$), with $\epsilon_{\text{SAP}_{\text{LS}}}$ derived from all the data and not by individual groups.


\subsection{Final set of HIs}
The full set of HIs produced by I-GLIDE are then aggregated with our UQ as the set
$\text{HI}_{\text{groups}} = \{ \epsilon_{\text{SAP}^{(g)}}, \epsilon_{\text{NAP}^{(g)}}, \epsilon_{\text{SAP}_{\textbf{LS}}}, \epsilon_{\text{NAP}_{\textbf{LS}}}, \sigma_{a^{(g)}}, \sigma_{e^{(g)}} \} \forall g \in G$
 where $\sigma_{a^{(g)}}, \sigma_{e^{(g)}}$ are respectively the aleatoric and espitemic uncertainties computed for each group g. 
We also compare with the monilithic architecture where the inputs $x$ are not divided into subgroups, and thus our set of HIs is defined as $\text{HI}_{\text{mono}} = \{ \epsilon_{\text{SAP}}, \epsilon_{\text{NAP}}, \epsilon_{\text{SAP}_{\textbf{LS}}}, \epsilon_{\text{NAP}_{\textbf{LS}}}, \sigma_{a}, \sigma_{e} \}$.

To evaluate their predictive capabilities, we train a meta regressor $\mathcal{F}(.)$ on the task of RUL estimation. We also compare with the previous set of RaPP indicators from Gonzàlez by define $\text{HI}_\text{González} = \{\epsilon_{\text{NAP}_\text{LS}}, \epsilon_{\text{SAP}_\text{LS}}\}$ \cite{González-Muñiz_Díaz_Cuadrado_García-Pérez_2022}.




\section{Experiments}
We aim to find out whether augmenting latent space HIs (RaPP metrics) with UQ, contributes to better understanding of degradation mechanisms in complex system in the perspective of RUL estimation. In a second step, we'd like to test whether introducing an architecture able to disentangle sub-systems degradation mechanisms can further improve this RUL estimation, through better understanding of the system from data, and minimal domain-knowledge. Our code is available at: https://github.com/LucasStill/I-GLIDE for reproduction purposes.


\begin{table}[t]
\caption{The four C-MAPSS dataset subsets and their description, with associated number of operating conditions and amount of degradation fault modes origins.}
\centering
\begin{tabular}{|c|c|c|c|c|}
\hline
 & \textbf{FD001} & \textbf{FD002} & \textbf{FD003} & \textbf{FD004} \\ \hline
\textbf{Train trajectories} & 100 & 260 & 100 & 248 \\ \hline
\textbf{Test trajectories} & 100 & 259 & 100 & 249 \\ \hline
\textbf{Conditions} & 1 & 6 & 1 & 6 \\ \hline
\textbf{Fault modes} & 1 & 1 & 2 & 2 \\ \hline
\end{tabular}
\label{tab:dataset_details}
\end{table}

\subsection{Datasets}
We evaluate our framework on two datasets. The C-MAPSS dataset \cite{Saxena_Goebel_Simon_Eklund_2008} a benchmark for degradation modeling, contains simulated run-to-failure trajectories of jet engines generated using NASA’s C-MAPSS simulator. Each multivariate time series corresponds to a unique engine operating under varying conditions, divided into four subsets (FD001–FD004; see Table \ref{tab:dataset_details}). During training, we focus on samples with RUL $\leq$ 80 timesteps ($R_{early}=80$) to prioritize early degradation signals while retaining healthy-state representations. For testing, we follow the established protocol by prior works with $R_{early}=125$ to enable direct comparison. The test set contains truncated trajectories that stop before the point of failure, and the task is to predict this last available value on the trajectory. We present the partitioning of the different groups in table \ref{tab:sensor_grouping}.

To further validate our approach and compare the effects of subsystem group separation, we test on the MILL NASA dataset, which records 167 unique tool wear progression during milling experiments under varied conditions (depth of cut, feed rate, material). Sensor signals (acoustic, vibration, current) track wear, with failure defined at wear=0.70 (initial wear=0). We classify samples as healthy (wear $\leq$ 0.20) during training and evaluate degradation over three test phases: complete trajectories, moderate degradation (wear > 0.20), and severe degradation (wear > 0.50). Each sensor contains a total of 9000 entries, but some have missing data which we fill through interpolation with neighboring values.

For both datasets, we will apply our method to create our sets of HIs and measure their predictive capabilities over RUL prediction, using RMSE metric which is commonly used on these datasets \cite{De_Pater_Mitici_2022}.

\begin{table}[t]
\centering
\caption{Grouping of sensors in the CMAPS dataset.}
\begin{tabular}{lll}
\toprule
\textbf{Group} & \textbf{Sensor ID} & \textbf{Description} \\
\midrule
\multirow{5}{*}{Fan}
    & s\_1 & Total temperature at fan inlet (°R) \\
    & s\_5 & Pressure at fan inlet (psia) \\
    & s\_8 & Physical fan speed (rpm) \\
    & s\_13 & Corrected fan speed (rpm) \\
    & s\_18 & Demanded fan speed (rpm) \\
    & s\_19 & Demanded corrected fan speed (rpm) \\
\midrule
LPC
    & s\_2 & Total temperature at LPC outlet (°R) \\
\midrule
\multirow{3}{*}{HPC}
    & s\_3 & Total temperature at HPC outlet (°R) \\
    & s\_7 & Total pressure at HPC outlet (psia) \\
    & s\_11 & Static pressure at HPC outlet (psia) \\
\midrule
\multirow{2}{*}{Core}
    & s\_9 & Physical core speed (rpm) \\
    & s\_14 & Corrected core speed (rpm) \\
\midrule
\multirow{3}{*}{Pressure Turbine}
    & s\_4 & Total temperature at LPT outlet (°R) \\
    & s\_20 & HPT coolant bleed (lbm/s) \\
    & s\_21 & LPT coolant bleed (lbm/s) \\
\midrule
\multirow{6}{*}{Other}
    & s\_6 & Total pressure in bypass-duct (psia) \\
    & s\_10 & Engine pressure ratio (P50/P2) (-) \\
    & s\_12 & Ratio of fuel flow to Ps30 (pps/psia) \\
    & s\_15 & Bypass Ratio (-) \\
    & s\_16 & Burner fuel-air ratio (-) \\
    & s\_17 & Bleed Enthalpy (-) \\
\bottomrule
\end{tabular}
\label{tab:sensor_grouping}
\end{table}

\subsection{Experimental Methodology}


To validate our framework, we adopt a prognostics-centric evaluation protocol that directly benchmarks HIs by their ability to predict RUL—the ultimate objective of HI construction. We first compare the RaPP-based HIs proposed by González et al. ($\text{HI}_\text{González}$) \cite{González-Muñiz_Díaz_Cuadrado_García-Pérez_2022} against a enhanced variants: $\text{HI}_\text{mono}$, which integrates encoder-level RaPP metrics \cite{Ki_Hyun_Sangwoo_Yongsub}, with UQ. We use a single timestep for each created HI. Critically, we bypass classical HI metrics like monotonicity or trendability, which often fail to correlate with actionable prognostic value, and instead train a random forest (RF) regressor $\mathcal{F}$ to map HIs to RUL. This choice reflects a key design principle: HI quality should be first judged by its downstream utility in prognostics. 

We then introduce our $\text{I-GLIDE}$ architecture, instantiated as $\text{I-GLIDE}_\text{AE}$ and $\text{I-GLIDE}_\text{VAE}$, which generates subsystem-specific HIs ($\text{HI}_\text{groups}$) by isolating sensor-group degradation patterns. These are benchmarked against monolithic AE/VAE counterparts under identical RF training protocols, ensuring fair comparison. By using a simple, non-temporal model like RF, we deliberately decouple HI quality from algorithmic sophistication, isolating how architectural choices (monolithic vs. subsystem-specific) impact prognostics performance.

\subsection{Results}

When comparing the RUL estimation capabilities of $\text{HI}_\text{González}$ and $\text{HI}_\text{mono}$ as shown in table \ref{tab:hi_rul2}, we find that $\text{HI}_\text{mono}$ consistently outperforms $\text{HI}_\text{González}$ across all C-MAPSS subsets (FD001-FD004). For example, $\text{HI}_\text{mono}$ derived from AEs reduces RMSE by 22.95\% on average compared to $\text{HI}_\text{González}$, with particularly notable gain on FD002 (15.71 vs 22.91) and FD003 (8.07 vs 12.03). Similar trends hold for VAEs, where $\text{HI}_\text{mono}$ achieves a 28.44\% average RMSE improvement, underscoring the value of UQ in stabilizing HI quality. The same can be observed for the MILL dataset in table \ref{tab:mill_scores}. This ablation study demonstrates that a broader coverage of latent HIs with UQ, collectively strenghtens RUL predictive capabilities, even before subsytem-specific modeling. 

Next, we deploy $\text{I-GLIDE}$, which explicitly disentangles subsytem degradation (e.g., HPC, fan, turbine in C-MAPSS) by grouping sensor signals into functionally coherent components (exacts group choices are presented in the appendix). Compared to monolithic architectures, $\text{I-GLIDE}$ achieves superior robustness, as evidenced by its 39.96\% reduction of standard deviation in RMSE across C-MAPSS subsets from AE-based HIs (\ref{tab:model_metrics2}). For VAEs, gains are even more pronounced: $\text{I-GLIDE}_\text{VAE}$ reduces RMSE by 39.03\% and standard deviation by 56.07\%, resolving the instability seen in monolithic VAEs (e.g., FD002/FD003 variance). This subsystem isolation proves critical on FD004-the most complex C-MAPSS subset-where $\text{I-GLIDE}$'s average results set a new state-of-the-art performance with a RMSE of 14.19 despite using only a RF regressor for RUL prediction. 

On the MILL dataset, $\text{I-GLIDE}$'s subsytem-specific HIs improve RUL prediction across all degradation phases (healthy, moderate, severe), with $\text{I-GLIDE}_\text{AE}$-driven HIs achieving the lowest RMSE in every scenario. VAE gains are subtler, likely due to MILL's lower inherent complexity, or high dimensional space which limits the benefits of variational inference.

\begin{table}[h]
\centering
\caption{Comparison of sets of HIs extracted from different architectures to predict the RUL RMSE on C-MAPSS test dataset using a Random Forest for $\mathcal{F}$. Best models shown.}
\begin{tabular}{|l|l|c|c|c|c|c|}
\hline
\textbf{HI Extractor} & \textbf{HI Set for $\mathcal{F}(.)$} & \textbf{FD001} & \textbf{FD002} & \textbf{FD003} & \textbf{FD004} & \textbf{Avg.} \\
\hline 
\multirow{2}{*}{AE} & $\text{HI}_\text{González}$ \cite{González-Muñiz_Díaz_Cuadrado_García-Pérez_2022} & 11.43 & 22.91 & 12.03 & 16.78 & 15.79 \\
                     & $\text{HI}_\text{mono}$ & 10.53 & \textbf{15.71} & \textbf{8.07} & 14.35 & 12.17 \\ \hline \hline
\multirow{2}{*}{VAE} & $\text{HI}_\text{González}$ \cite{González-Muñiz_Díaz_Cuadrado_García-Pérez_2022} & 27.56 & 28.62 & 24.36 & 22.33 & 25.72 \\
                      & $\text{HI}_\text{mono}$ & 18.77 & 19.44 & 15.59 & 19.81 & 18.40 \\ \hline \hline
$\text{I-GLIDE}_\text{AE}$ & $\text{HI}_\text{groups}$ & \textbf{9.47} & 16.18 & 8.29 & 12.32 & 11.57 \\ 
$\text{I-GLIDE}_\text{VAE}$ & $\text{HI}_\text{groups}$ & 12.33 & 16.76 & 8.5 & \textbf{11.4} & 12.25 \\
\hline
\end{tabular}
\label{tab:hi_rul2}
\end{table}

\begin{table}[h]
\centering
\caption{RUL MILL Dataset Benchmark on three wear levels. I-GLIDE HIs consistently outperform the monolithic counterpart in RMSE for RUL prediction \cite{mill_nasa}.}
\begin{tabular}{|c|c|c|c|}
\hline
\textbf{Model Name, \text{HI} set for RF} & \textbf{Wear 0.0-0.70} & \textbf{Wear 0.20-0.70} & \textbf{Wear 0.50-0.70} \\
\hline
AE, $\text{HI}_\text{González}$ \cite{González-Muñiz_Díaz_Cuadrado_García-Pérez_2022} & 23.78 & 24.34 & 22.33 \\ \hline
AE, $\text{HI}_\text{mono}$ & 16.14 & 16.47 & 16.25 \\ \hline
$\text{I-GLIDE}_\text{AE}$, $\text{HI}_\text{groups}$ & \textbf{13.64} & \textbf{14.37} & \textbf{16.17} \\ \hline\hline
VAE, $\text{HI}_\text{González}$ \cite{González-Muñiz_Díaz_Cuadrado_García-Pérez_2022} & 27.84 & 27.92 & 27.14 \\ \hline
VAE, $\text{HI}_\text{mono}$ & 22.32 & 22.46 & 23.47 \\ \hline
$\text{I-GLIDE}_\text{VAE}$, $\text{HI}_\text{groups}$ & \textbf{21.76} & \textbf{22.29} & \textbf{23.13} \\ \hline

\end{tabular}
\label{tab:mill_scores}
\end{table}

\begin{table}[h]
\centering
\caption{Average model performances across 10 runs over C-MAPSS subsets using RMSE (mean ± standard deviation). Bold: best results per subset; underline: outperforms methods without HIs. Last column provides average improvement over the previous row.}
\label{tab:model_metrics2}
\begin{tabular}{|l|c|c|c|c|c|c|}
\hline
\textbf{Model, HI Set} & \textbf{FD001} & \textbf{FD002} & \textbf{FD003} & \textbf{FD004} & \textbf{Avg.} & \makecell{\textbf{Improvement}} \\ \hline
AE, $\text{HI}_\text{González}$ \cite{González-Muñiz_Díaz_Cuadrado_García-Pérez_2022} &
\makecell{19.00 \\ ±4.78} & \makecell{25.69 \\ ±4.19} & \makecell{18.38 \\ ±6.18} & \makecell{19.46 \\ ±2.46} & \makecell{20.63 \\ ±4.40} & -- \\ \hline
AE, $\text{HI}_\text{mono}$ &
\makecell{13.14 \\ ±2.50} & \makecell{20.35 \\ ±3.46} & \makecell{13.87 \\ ±5.07} & \makecell{17.73 \\ ±3.56} & \makecell{16.27 \\ ±3.65} & \makecell{+21.13\% \\ +17.15\%} \\ \hline
$\text{I-GLIDE}_\text{AE}$, $\text{HI}_\text{groups}$ &
\makecell{\textbf{12.11} \\ ±2.72} & \makecell{22.01 \\ ±2.88} & \makecell{\textbf{10.23} \\ ±1.85} & \makecell{\underline{14.92} \\ ±1.31} & \makecell{14.82 \\ ±2.19} & \makecell{+8.94\% \\ +39.96\%} \\ \hline \hline
VAE, $\text{HI}_\text{González}$ \cite{González-Muñiz_Díaz_Cuadrado_García-Pérez_2022} &
\makecell{34.13 \\ ±3.71} & \makecell{31.05 \\ ±1.89} & \makecell{27.25 \\ ±2.58} & \makecell{25.23 \\ ±2.03} & \makecell{29.42 \\ ±2.55} & -- \\ \hline
VAE, $\text{HI}_\text{mono}$ &
\makecell{27.19 \\ ±5.97} & \makecell{22.81 \\ ±2.86} & \makecell{24.64 \\ ±5.26} & \makecell{22.89 \\ ±1.82} & \makecell{24.38 \\ ±3.98} & \makecell{+17.10\% \\ -55.83\%} \\ \hline
$\text{I-GLIDE}_\text{VAE}$, $\text{HI}_\text{groups}$ &
\makecell{15.32 \\ ±2.08} & \makecell{\textbf{18.83} \\ ±1.51} & \makecell{11.12 \\ ±2.29} & \makecell{\textbf{\underline{14.19}} \\ ±1.11} & \makecell{14.87 \\ ±1.75} & \makecell{+39.03\% \\ +56.07\%} \\ \hline
\end{tabular}
\end{table}

\section{Discussion}

Remarkably, when looking at the best produced models, even with a RF-a model far simpler than deep learning baselines-our HIs match or exceed prior SOTA on three out of four C-MAPSS benchmarks as shown in table \ref{tab:direct_rul}. This paradox highlights that HI quality, not model complexity, drives prognostics success. When looking at the expected accuracies of the different models, we see that $\text{I-GLIDE}$ has lower standard deviations, being more robust to prediction, which explains why in two cases the best model was a monolithic AE: despite showing great performance on a single set, its high standard deviation shown in table \ref{tab:model_metrics2} indicates it would not be robust on a broader test set, or in real-life conditions. This is why $\text{I-GLIDE}$ offers solid perspectives towards more robust predictions. 

\begin{table}[h]
\centering
\caption{Comparison of I-GLIDE method for HI extraction benchmarked to predict a RUL, compared with best known approaches. In bold are the best results for each subset. Most previous methods were predicting a RUL from transformed sensor data without producing HIs, contrarily to our method which does provide HIs.} 
\begin{tabular}{|l|c|c|c|c|}
\hline
\textbf{Model} & \textbf{FD001} & \textbf{FD002} & \textbf{FD003} & \textbf{FD004} \\
\hline
MLP \cite{Zheng_Ristovski_Farahat_Gupta_2017} & 37.56 & 80.03 & 37.39 & 77.37 \\ \hline
CNN \cite{Zheng_Ristovski_Farahat_Gupta_2017} & 18.45 & 30.29 & 19.82 & 29.16 \\ \hline
CNN-LSTM \cite{Peng_Chen_Chen_Tang_Li_Gui_2021} & 11.17 & - & 9.99 & - \\ \hline
MS-DCNN \cite{Li_Zhao_Zhang_Zio_2020} & 11.44 & 19.35 & 11.67 & 22.22 \\ \hline
VAE + RNN \cite{Costa_Sánchez_2022} & 11.44 & 24.12 & 14.88 & 26.54 \\ \hline
MLE(4X)+CCF \cite{Pillai_Vadakkepat_2021} & 11.57 & 18.84 & 11.83 & 20.78 \\ \hline
RVE \cite{Costa_Sánchez_2022} & 13.42 & 14.92 & 12.51 & 16.37 \\ \hline
Probabilistic RUL CNN \cite{De_Pater_Mitici_2022} & 12.42 & \textbf{13.72} & 12.16 & 15.95 \\ \hline 
$\text{I-GLIDE}_\text{AE}$ + RF (ours) & \textbf{9.47} & 16.18 & \textbf{8.29} & 12.32 \\ \hline
$\text{I-GLIDE}_\text{VAE}$ + RF (ours) & 12.33 & 16.76 & 8.5 & \textbf{11.4} \\ \hline
\end{tabular}
\label{tab:direct_rul}
\end{table}

\begin{figure}[h]
\centering
\subfigure[Latent Encoder HI]{\includegraphics[width=0.32\textwidth]{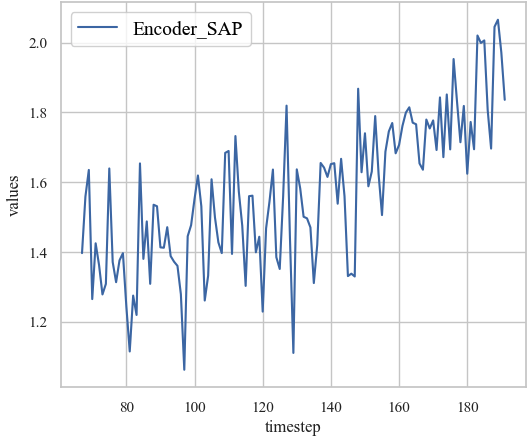}}
\hfill
\subfigure[Latent 
$z$ HIs]{\includegraphics[width=0.32\textwidth]{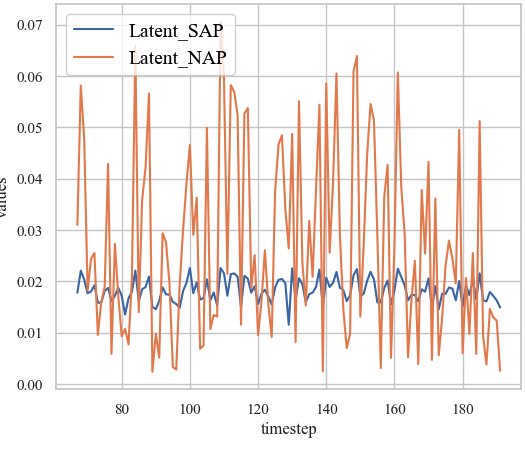}}
\hfill
\subfigure[Epistemic UQ HI]{\includegraphics[width=0.32\textwidth]{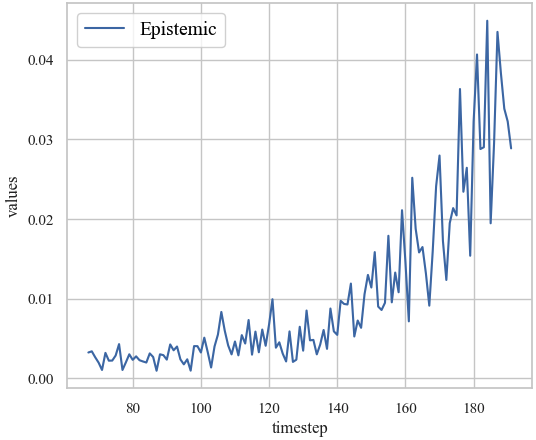}}
\label{fig:FD001_AE_base_HIs}
\caption{AE HI trajectories for Engine 1 for the monolithic architecture. We can observe that the HIs model a degradation, but cannot distinguish sub-system components. We only show the SAP metric for the encoder HIs because NAP shows extreme values.}
\end{figure}

\begin{figure}[h]
\centering
\subfigure[Latent Encoder HIs]{\includegraphics[width=0.32\textwidth]{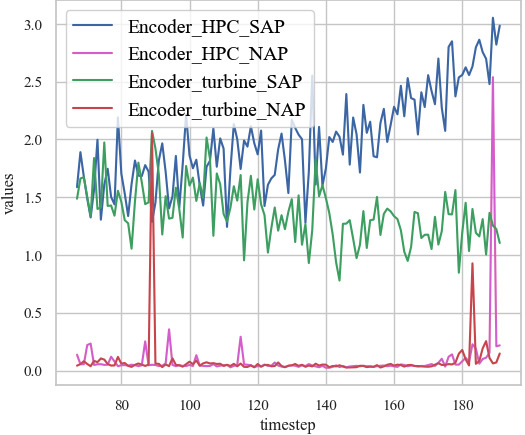}}
\hfill
\subfigure[Latent $z$ HIs]{\includegraphics[width=0.32\textwidth]{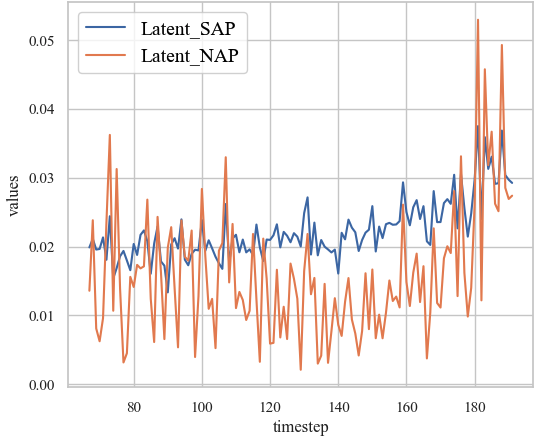}}
\hfill
\subfigure[Epistemic UQ HI]{\includegraphics[width=0.32\textwidth]{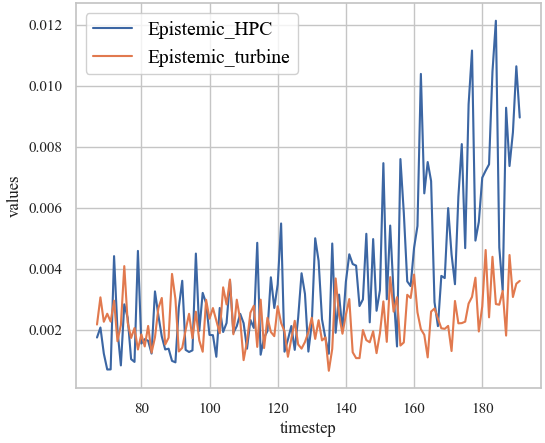}}
\label{fig:FD001_AE_groups_HIs}
\caption{$\text{I-GLIDE}_\text{AE}$ HI trajectories for Engine 1, comparing degradation effects on HPC and Turbine. Latent encoder HIs (a) show rising HPC degradation and reduced Turbine HIs due to cross-component effects. System-wide latent $z$ HIs trends are in (b). Epistemic uncertainty (c) rises sharply for HPC as degradation progresses, remaining stable for the Turbine until late-cycle HPC interference. UQ confirms causal cross-component effects without confusing intrinsic health states.}
\end{figure}

Monolithic AEs struggle to disentangle subsystem-specific degradation, which we illustrate with a HI plot of the trajectories in Figure \ref{fig:FD001_AE_base_HIs}.
For Engine 1 (FD001), where HPC degradation is the source, $\text{HI}_{\text{mono}}$ shows weak latent-space ($z$) sensitivity to subsystem dynamics. This occurs because deeper layers in monolithic AEs compress sensor signals into a global representation, obscuring non-stationary interactions (e.g., HPC wear indirectly altering turbine behavior). In contrast, Figure \ref{fig:FD001_AE_groups_HIs} reveals how I-GLIDE isolates these dynamics: the HPC encoder HI exhibits a clear upward trend, while the turbine HI shifts abruptly as degradation propagates—a causal linkage masked in monolithic architectures. Notably, the shared latent $z$ in I-GLIDE still captures the composite degradation trend, and subsystem-specific decoders also localize fault origins (e.g., rising epistemic uncertainty in HPC vs. stable turbine estimates). This explains why $\text{HI}_{\text{mono}}$ underperforms—it conflates cross-subsystem effects into a single noisy signal, while our I-GLIDE overcomes these restrains. In future work, we would like to formalize methods to interpret such causal relationships between HIs, identify noise patterns in the degradation signals, and apply it to maintenance tasks.

Traditional HI metrics (monotonicity, trendability, prognosability) often produce misleading scores (e.g., near-perfect prognosability) that poorly correlate with actual RUL prediction. Worse, they ignore subsystem-specific degradation, obscuring actionable insights. Our framework addresses this by directly linking HI quality to RUL prediction accuracy—a metric aligned with real-world decision-making. By disentangling subsystem trends (e.g., turbine wear vs. fan imbalance), I-GLIDE enables targeted fault diagnosis and maintenance planning. 

While our framework advances subsystem-aware HIs, several constraints merit consideration. First, it is worth noting that both the C-MAPSS and MILL datasets model exponential degradation patterns, which oversimplify real-world scenarios where industrial systems often exhibit linear or piecewise degradation trends. Real-world applications also introduce complex noise profiles (e.g., cyclic sensor artifacts) and heterogeneous failure modes that our method may not optimally capture without tailored adaptations.

Readers should be aware that our architecture assumes strictly monotonous degradation, limiting its ability to model recovery phases—a critical shortcoming for systems where transient improvements occur, such as medical devices supporting patient recovery or aircraft exiting high-stress environments. Furthermore, our subsystem groupings rely on domain heuristics; while this aligns with prior work, poorly defined sensor groupings could propagate biases into the latent representations, undermining HI interpretability.

\section{Conclusion and Future Work}

This work establishes the first prognostics benchmark for evaluating Health Indicators (HIs) generated via the RaPP methods, demonstrating that integrating uncertainty quantification significantly enhances their predictive capabilities. Building on this foundation, we introduce I-GLIDE, a novel framework that learns subsystem-specific latent representations through dedicated encoder-decoder pairs. By isolating degradation mechanisms (e.g., HPC degradation vs. turbine wear) while maintaining global system dynamics via a shared latent space, I-GLIDE captures nuanced failure modes without compromising system-level coherence. The resulting high-quality HIs achieve state-of-the-art performance on the C-MAPSS dataset, surpassing existing deep learning benchmarks using only a simple Random Forest regressor.

Our subsystem-specific HIs advance prognostics but invite refinement. Temporal improvements—like extending observation windows—could better resolve slow degradation signatures and transient noise, aligning HI trajectories with real-world failure timelines. Coupling uncertainty-specific t-SNE visualizations with expert annotations could map latent clusters to physical degradation stages, bridging data-driven insights with domain knowledge.

A promising direction involves modeling causal subsystem interactions via architectures like graph neural networks, trained on fused HIs to disentangle degradation propagation (e.g., turbine-to-compressor wear). This would scale prognostics to systems with complex interdependencies.

Critically, our results show that high-quality HIs paired with simple models (e.g., RF) outperform deep learning on raw data—a "data is gold" paradigm. Future efforts should prioritize refining physics-aware HI representations—grounded in subsystem dynamics and enriched with UQ to unlock generalizable, trustworthy RUL prediction across grounded industrial domains.

\begin{credits}
\subsubsection{\ackname} This work has been supported by the French government under the "France 2030” program, as part of the SystemX Technological Research Institute within the JNI3 project.

\subsubsection{\discintname}
The authors have no competing interests to declare that are relevant to the content of this article.
\end{credits}

%
%
%
\bibliographystyle{splncs04}

\newpage

\section{Appendix}

\begin{figure}[h]
\centering
\subfigure[SAP Encoder HIs]{\includegraphics[width=0.45\textwidth]{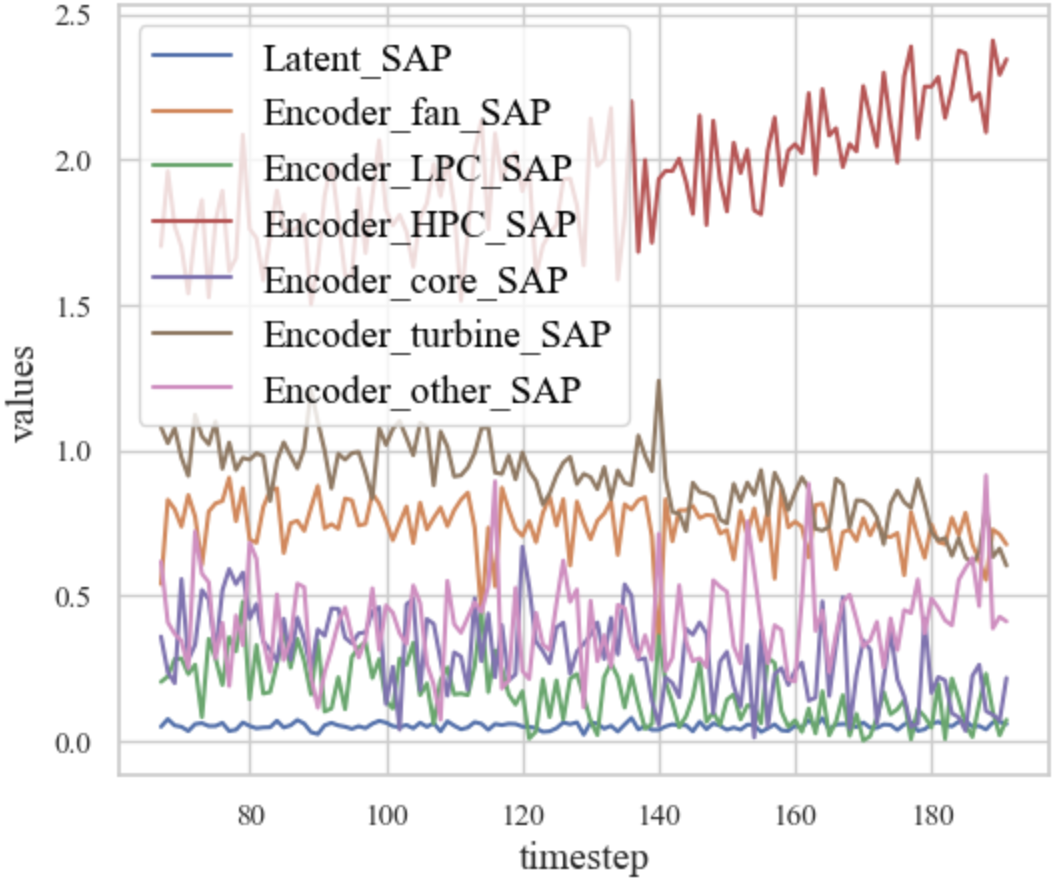}}
\hfill
\subfigure[NAP Encoder HIs]{\includegraphics[width=0.45\textwidth]{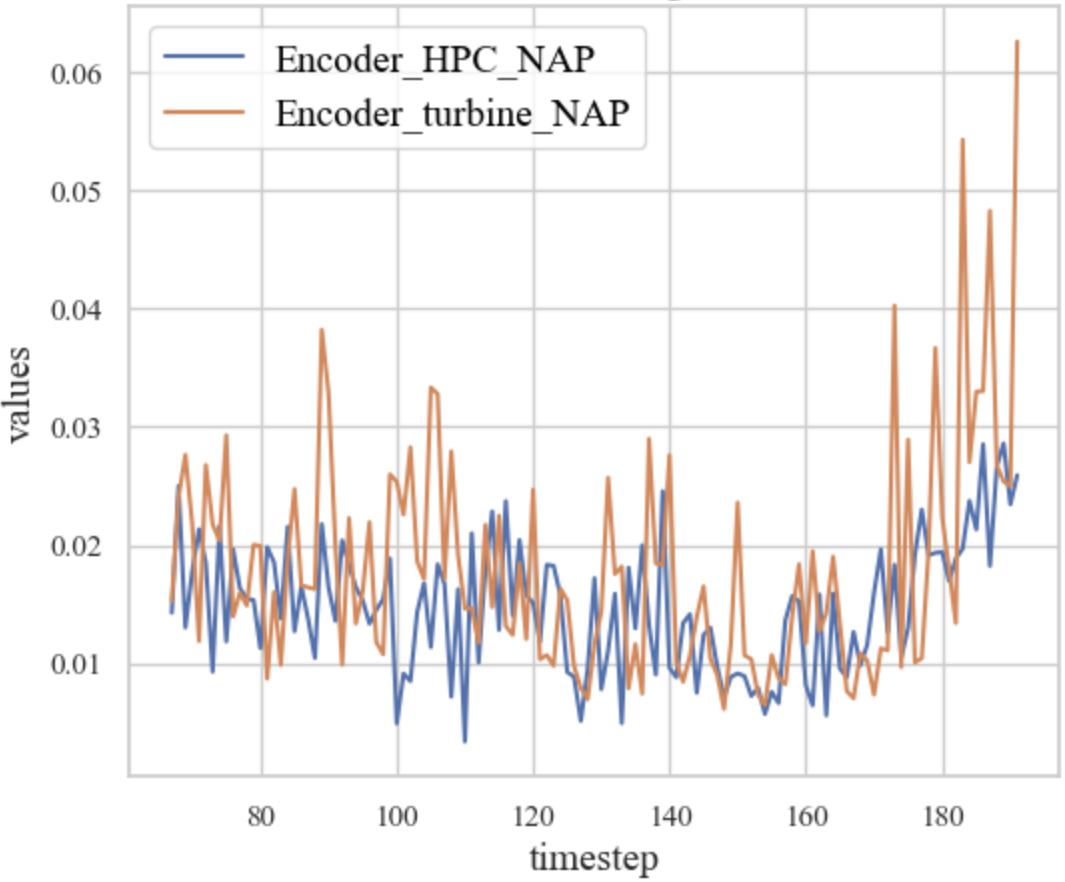}}
\vspace{1em} 
\subfigure[Latent $z$ RaPP HIs]{\includegraphics[width=0.45\textwidth]{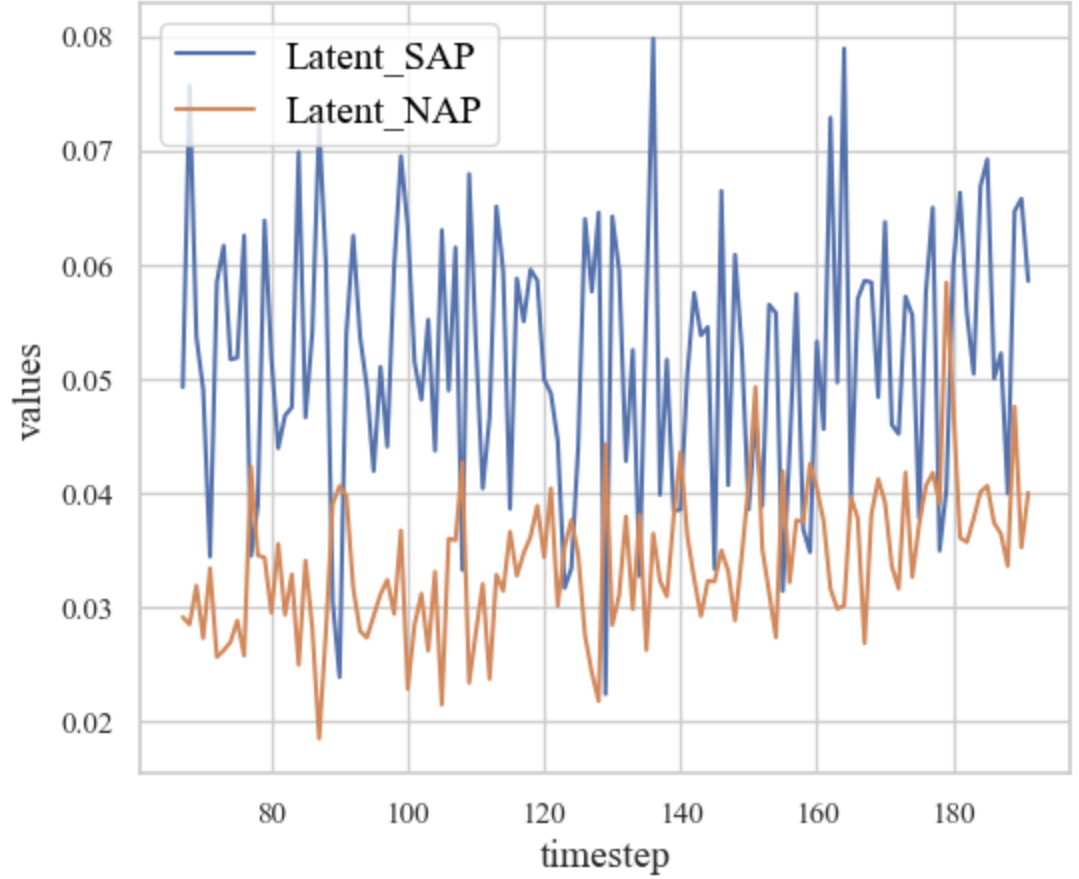}}
\hfill
\subfigure[Aleatoric and Epistemic UQ HI]{\includegraphics[width=0.45\textwidth]{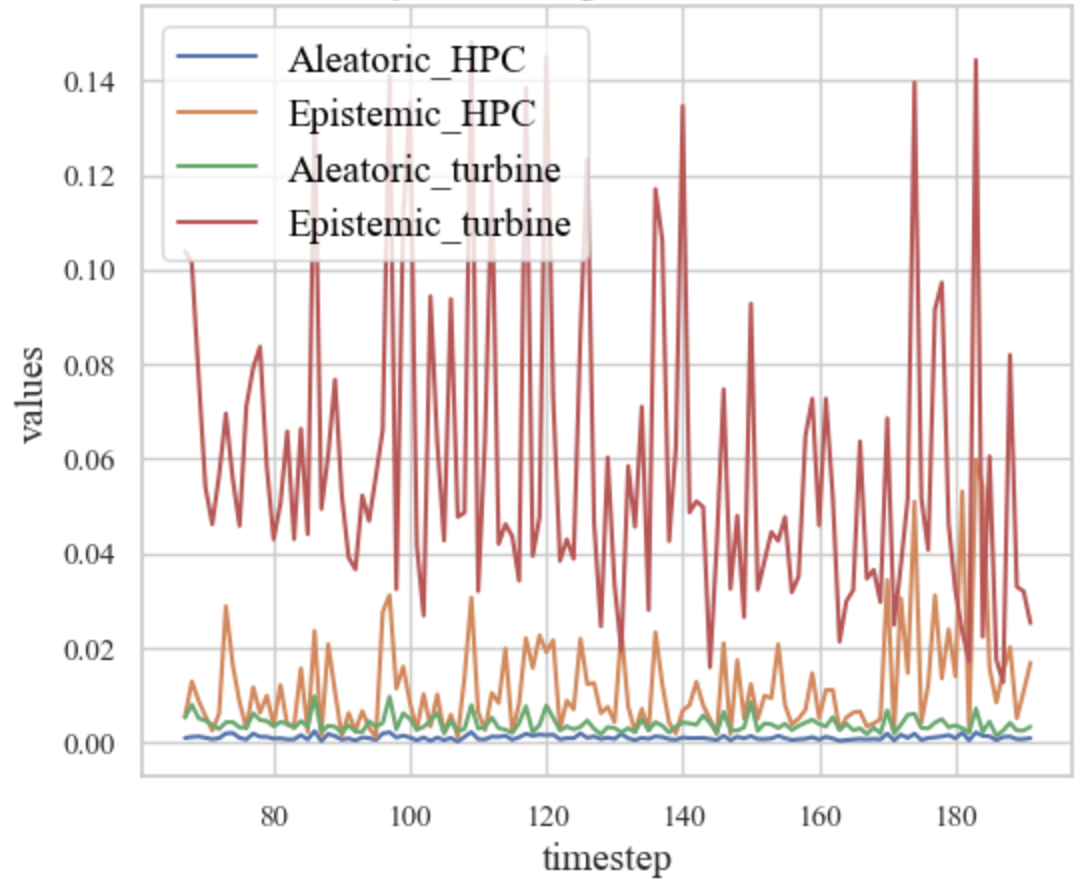}}
\label{fig:FD001_VAE_groups_HIs}
\caption{$\text{I-GLIDE}_\text{VAE}$ HI trajectories for Engine 1, comparing degradation effects of all sub-system. Latent encoder SAP HIs (a) show rising HPC degradation and a decreased of HIs from other sub-systems. In b) we see that both HIs seem to follow a system-wide latent $z$ HIs trends are in (c). 
Epistemic uncertainty (d) We show the VAE uncertainties for two sub-systems: HPC and turbine. We observe that the turbine value has an overall decreasing trend whereas HPC increases near the end. The uncertainties values are not normalized between groups which can be a hindrance, that is why a meta regressor $\mathcal{F}$ can learn how to interpret them automatically..}
\end{figure}

\begin{table}[h!]
\centering
\caption{Hyperparameters used for training the Random Forest model.}
\begin{tabular}{ll}
\toprule
\textbf{Hyperparameter} & \textbf{Value} \\
\midrule
\texttt{n\_estimators} & 100 \\
\texttt{max\_depth} & 10 \\
\texttt{random\_state} & 42 \\
\texttt{min\_samples\_split} & 2 \\
\bottomrule
\end{tabular}
\label{tab:rf_hyperparameters}
\end{table}

\begin{table}[h!]
\centering
\caption{Hyperparameters and settings used for training the Autoencoder model.}
\begin{tabular}{ll}
\toprule
\textbf{Parameter/Setting} & \textbf{Value} \\
\midrule
\texttt{window\_size} & 1 \\
\texttt{test\_size} & 0.3 \\
\texttt{scaler} & MinMaxScaler \\
\texttt{optimizer} & Adam \\
\texttt{criterion} & Mean Squared Error (MSELoss) \\
\texttt{batch\_size} & 128 \\
\texttt{epochs} & 200 \\
\bottomrule
\end{tabular}
\label{tab:autoencoder_hyperparameters}
\end{table}

\begin{table}[h!]
\caption{Architecture of the Monolithic Autoencoder model on CMAPS.}
\centering
\begin{tabular}{>{\raggedright\arraybackslash}p{2cm} >{\raggedright\arraybackslash}p{8cm}}
\toprule
\textbf{Component} & \textbf{Architecture} \\
\midrule
Encoder & Linear(input\_dim, 10) $\rightarrow$ ReLU $\rightarrow$ Linear(10, 20) $\rightarrow$ ReLU $\rightarrow$ Linear(20, 10) $\rightarrow$ ReLU \\
\midrule
Latent Layer & Linear(10, 2) \\
\midrule
Decoder & Linear(2, 10) $\rightarrow$ Dropout(0.2) $\rightarrow$ ReLU $\rightarrow$ Linear(10, 20) $\rightarrow$ Dropout(0.2) $\rightarrow$ ReLU $\rightarrow$ Linear(20, 10) $\rightarrow$ Dropout(0.2) $\rightarrow$ ReLU $\rightarrow$ Linear(10, input\_dim) \\
\bottomrule
\end{tabular}
\label{tab:autoencoder_architecture}
\end{table}

\begin{table}[h!]
\centering
\caption{$\text{I-GLIDE}_\text{AE}$ Architecture. Note that due to the introduction of sub-system specific groups, and the fact they have varying sizes, the bottleneck effect learned at each subsequent layer of the encoder is affected. As we wanted to retain an overall architecture similar to the vanilla AE, we decided to allow this sparsity to remain notably in the second layer.}
\begin{tabular}{>{\raggedright\arraybackslash}p{2cm} >{\raggedright\arraybackslash}p{8cm}}
\toprule
\textbf{Component} & \textbf{Architecture} \\
\midrule
Encoder (for each group) & Linear(input\_group\_dim, 10) $\rightarrow$ ReLU $\rightarrow$ Linear(10, 20) $\rightarrow$ ReLU $\rightarrow$ Linear(20, 10) $\rightarrow$ ReLU \\
\midrule
Latent Layer & Linear(10 $\times$ number\_of\_groups, latent\_dim) \\
\midrule
Decoder (for each group) & Linear(latent\_dim, 10) $\rightarrow$ Dropout(0.2) $\rightarrow$ ReLU $\rightarrow$ Linear(10, 20) $\rightarrow$  Dropout(0.2) $\rightarrow$ ReLU $\rightarrow$ Linear(20, 10) $\rightarrow$ Dropout(0.2) $\rightarrow$ ReLU $\rightarrow$ Linear(10, input\_group\_dim) \\
\bottomrule
\end{tabular}
\label{tab:multi_autoencoder_architecture}
\end{table}

\subsection{Uncertainty Quantification}

\subsubsection{Bayesian Frameworks for Uncertainty}  
In Bayesian deep learning, model parameters $W_E$ and latent variables $z$ are treated probabilistically. For regression tasks, the likelihood is modeled as a Gaussian:  
\begin{equation}  
p(y | W, x) = \mathcal{N}\left(y; \, \mu^{W_E}(x), \, \sigma^{W_E}(x)^2\right),  
\label{eq:reg_likelihood}  
\end{equation}  
where $\mu^{W_E}(x)$ and $\sigma^{W_E}(x)$ are encoder-derived mean and variance. Monte Carlo (MC) dropout \cite{gal2016dropoutbayesianapproximationrepresenting,Hinton_Srivastava_Krizhevsky_Sutskever_Salakhutdinov_2012} approximates Bayesian inference by sampling from the posterior 
via stochastic dropout masks during inference. 
The predictive distribution integrates over both \( W \) and latent variables \( z \):  
\begin{equation}  
p(y | x) = \int \mathcal{N}\left(y; \, \mu^{W_{E}}(x, z), \, \sigma^{W_{E}}(x, z)^2\right) p(W | \mathcal{X}) p(z | x) \, dW_{E} dz,  
\label{eq:reg_predictive1}  
\end{equation}  
where \( \mathcal{X} \) is the training data. This intractable integral is approximated via MC sampling:  
\begin{equation}  
p(y | x) \approx \frac{1}{n} \sum p(y|\hat{W}_E, x, \hat{z}),  
\label{eq:reg_predictive2}  
\end{equation}  
with $\hat{W}_E \sim p(W_E)$ (via dropout) and $\hat{z} \sim p(z | x)$.  

\subsubsection{Variational Autoencoders for Aleatoric Uncertainty}  
Variational autoencoders (VAEs) \cite{Kingma_Welling_2013} explicitly model aleatoric uncertainty through a stochastic latent space $\hat{z} \sim q(z|x)$, where $q(z|x)$ is the encoder’s approximate posterior. The training objective combines reconstruction loss and KL divergence regularization:  
\begin{equation}  
\mathcal{L}_{\text{VAE}} = -\mathbb{E}_{q(z|x)}\left[\log p(x|z)\right] + \beta \cdot D_{\text{KL}}\left(q(z|x) \parallel p(z)\right),  
\end{equation}  
with $p(z)$ as a standard Gaussian prior. During inference:  
1. Aleatoric uncertainty is sampled via $\hat{z} \sim q(z|x)$.  
2. Epistemic uncertainty is introduced through dropout on decoder weights $W_E$.  

In the case of vanilla AEs, aleatoric uncertainty \( \sigma_a \) cannot be isolated as \( z \) is deterministic, rendering \( \sigma_a \) undefined. This highlights the advantage of VAEs for joint uncertainty estimation.

\begin{figure}[t]

\label{cmapss_schema}

\includegraphics[width=\textwidth]{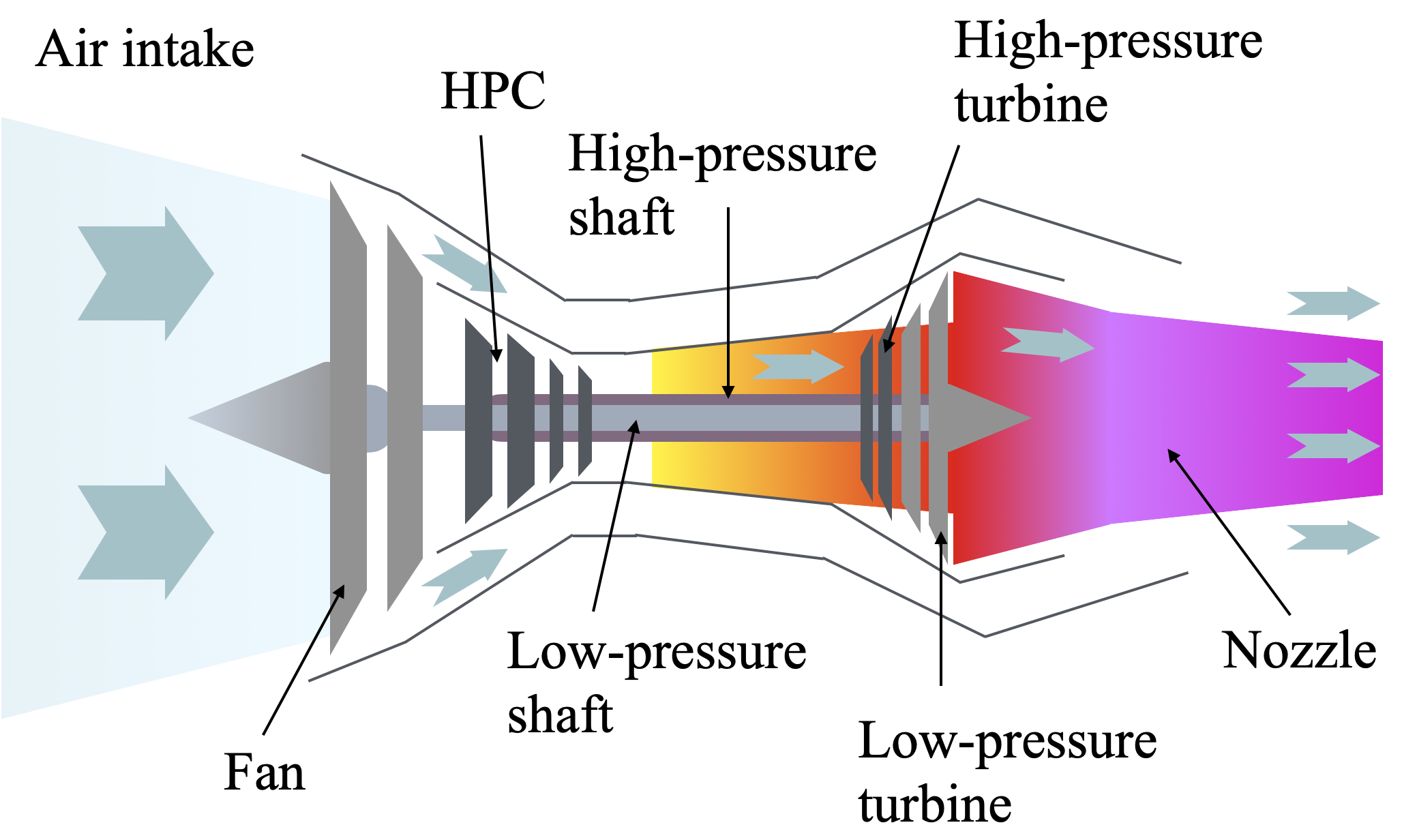}
\caption{\textbf{Turbofan Engine Schematic} –  
showing the different sub-systems that are represented in the C-MAPSS dataset \cite{Saxena_Goebel_Simon_Eklund_2008}.}  
\end{figure}

\end{document}